\documentclass[10pt,a4paper]{article}
\usepackage[T1]{fontenc}
\usepackage[utf8]{inputenc}
\usepackage{lmodern}
\usepackage[margin=0.87in]{geometry}
\usepackage{amsmath,amssymb,booktabs,array,graphicx,longtable}
\usepackage[hidelinks]{hyperref}
\usepackage{microtype}
\usepackage{enumitem}
\usepackage{needspace}

\setlist{nosep,leftmargin=*}
\title{Event-Only Wingbeat Counting under Camera Motion:\\A Controlled MuJoCo Benchmark}
\author{Zhang Nengbo\\[3pt]
\small School of Aerospace Engineering, Engineering Campus\\
\small Universiti Sains Malaysia, 14300 Nibong Tebal, Pulau Pinang, Malaysia\\
\small\texttt{zhangnb@student.usm.my}}
\date{September 2026}
\begin{document}
\maketitle
\begin{abstract}
Counting completed wingbeats requires identifying individual cycles, including during frequency changes and pauses; estimating a dominant frequency alone is insufficient. Camera motion further mixes target and background brightness changes in event observations. We present a controlled MuJoCo benchmark that separates motion training from event-only image translation compensation. The acquisition contains 324 streams from 24 independent scenes, three flapping geometries, two distances (1.5 and 3.0\,m), and static, moderate-motion and stronger-motion views. Fifteen scenes are used for fitting, three for validation and six for held-out testing. A fixed causal temporal convolutional network is evaluated in a matched $2\times2$ ablation with three initialization seeds and compared with ridge, Fourier, autocorrelation and an adapted EEPPR baseline. Under moderate motion, paired motion training reduces count mean absolute error from 31.130 to 3.185 cycles at 1.5\,m and from 42.019 to 5.444 at 3.0\,m. Adding the tested compensation increases these errors to 4.630 and 10.185, respectively. A Fourier baseline achieves 0.944 cycles at 1.5\,m under moderate motion, showing that the neural model is not uniformly best. We report exact-count accuracy and temporally matched cycle F1 alongside count error. These findings support motion-aware training in this small synthetic benchmark, while exposing limits of simple event-background stabilization. They do not establish real-sensor performance, aerodynamic flight, or generalization to unseen vehicle types.
\end{abstract}

\section{Introduction}
An observing micro aerial vehicle may need to determine how many times another vehicle has flapped its wings during an interval. The distinction between counting and frequency estimation matters: a frequency of 5\,Hz does not specify how many complete cycles occurred near the beginning or end of a recording, and a changing rate or pause invalidates a single frequency-times-duration estimate. A useful counter must identify individual completed cycles and retain its state as observations arrive.

Event cameras represent local brightness changes with signed, time-stamped events. Periodic event analysis has been studied through per-pixel filtering and spatiotemporal correlation \cite{frequency,eeppr}. For an observer facing a flapping vehicle, however, motion of the camera also produces events. This raises two separate questions: does exposure to motion during training improve counting, and does image stabilization estimated strictly from events provide an additional benefit? Their effects can be confounded if the training data, architecture and evaluation scenes change simultaneously.

We address these questions with a controlled simulation study rather than a new claim of universal state-of-the-art counting. The benchmark holds the temporal convolutional architecture and training budget fixed, crosses motion augmentation with causal background-based translation compensation, and keeps camera pose and inertial measurements outside inference. Baselines include a learned linear counter and causal period-to-count adaptations. Actual simulated wing-joint motion provides cycle labels, so the commanded frequency is not substituted for the count.

The contributions are: (i) a scene-separated, paired acquisition protocol with explicitly defined complete-cycle labels and near/far observations; (ii) a matched factorial comparison of motion training and strictly event-only compensation; and (iii) a transparent comparison against classical and event-specific periodic estimators, including startup penalties, unfavorable results and validation-selection failures. The independent test sample is six scenes. The intended contribution is therefore an auditable controlled benchmark and its conditional findings, not evidence of field deployment.

\section{Related work and relation to the earlier study}
\paragraph{Event-based periodic sensing.}
Frequency Cam uses asynchronous filtering and crossings to estimate pixel-wise periodicity \cite{frequency}. EEPPR estimates periodic rates by finding repeated spatiotemporal event patterns with three-dimensional correlation \cite{eeppr}. These methods motivate explicit periodic baselines, but periodic rate and completed-cycle counting are different outputs. Our EEPPR result includes a new causal phase reconstruction and counting adapter, a restricted spatial region, and a quantized time grid. It is not a reproduction of EEPPR's original frequency benchmark. Frequency Cam is discussed but was not implemented in this comparison; the baseline set is not exhaustive.

\paragraph{Repetition counting and temporal models.}
RepNet uses temporal self-similarity for class-agnostic repetition counting in videos \cite{repnet}. Temporal convolutions provide a general sequence-modeling approach \cite{tcn}. Here, a small causal temporal convolutional network (CNN) receives fixed spatial event summaries and predicts joint displacement. The study tests augmentation and compensation, not a new temporal architecture. Neither RepNet nor a full video transformer was trained on the current event dataset, and the results should not be read as a ranking of all repetition-counting methods.

\paragraph{Simulation and previous wingbeat work.}
MuJoCo provides the articulated-body simulation and renderer \cite{mujoco}. Event simulators such as ESIM and v2e address rendering and sensor realism \cite{esim,v2e}; our deterministic contrast sensor is simpler and is described explicitly below. This work follows an earlier RGB optical-flow study comparing convolutional, spiking and attention-based temporal models \cite{prior}. The present acquisition uses new scene seeds, high-rate simulated events, long continuous streams, complete-cycle labels and an event-only motion factorial experiment. The shared geometric setting and general counting motivation are inherited, but the earlier RGB results are not pooled with these data and are not directly comparable numerically. This paper does not report a new SNN--CNN--transformer experiment.

\section{Benchmark and counting protocol}
\subsection{Paired geometry and acquisition}
Three Crazyflie observer cameras face three adapted flapping geometries, identified by their upstream projects: RLFlapping, flappy\_v2 and BIRD\_SIM. Each camera targets its corresponding vehicle at a height of 1.3\,m, with camera-to-target-center distance 1.5\,m (near) or 3.0\,m (far), a $45^\circ$ vertical field of view and $128\times96$ pixels. Camera optics and resolution are held fixed across distances. The geometries have different sizes and articulated structures; they are not three recolorings of a single model. Sources and adaptations are recorded in Appendix~\ref{app:assets}.

The system uses inverse-dynamics pose and wing tracking as an optical observation test rig. Generalized support forces maintain the prescribed poses. Native MuJoCo physics and intensity rendering both run at 2,400\,Hz; supervision is recorded from actual joint positions and velocities. These procedures do not demonstrate autonomous hovering or validated flapping aerodynamics. Frequencies are test excitations, not asserted flight envelopes of the physical source vehicles.

The acquisition comprises 24 independent scene seeds. Scenes 0--14 are used for fitting, 15--17 for validation and 18--23 for testing. The development-to-test scene ratio is $18:6=3:1$; the fitting-to-test ratio alone is not 3:1. Each scene includes three targets, two distances, and static and moderate-motion views. The six held-out scenes additionally have a stronger-motion view. Thus there are 180 fitting streams, 36 validation streams and 108 test streams, for 324 total. Different target, distance and motion views from a scene remain in the same split.

Streams alternate between 12 and 16\,s. Profiles cycle through constant rate, chirp and stop--restart. Constant rates of 3, 5 and 7\,Hz are rotated across targets; the nonstationary schedules are specified in Appendix~\ref{app:schedule}. Backgrounds are balanced between plain and checker-textured native MuJoCo planes, including three of each among test scenes. Each of the three profiles occurs once per background in the test split. The two backgrounds use the same plane geometry, preventing a geometry change from being confounded with texture.

\subsection{Controlled camera motion}
For time $t$ in seconds and a scene-specific random phase $\phi$, moderate angular jitter is
\begin{align}
\theta_y(t)&=0.25\sin(2\pi\,0.7t+\phi)+0.05\sin(2\pi\,13t+\phi),\\
\theta_p(t)&=0.25\cos(2\pi\,0.7t+\phi)+0.05\cos(2\pi\,13t+\phi),
\end{align}
where angles are in degrees. The stronger condition multiplies both amplitudes by three and is held out from fitting and selection. The underlying wing trajectories are identical across distances and views of a scene. This intervention is small angular camera motion, not arbitrary six-degree-of-freedom flight or a target-tracking challenge. Camera commands and poses are used by acquisition only, never by the tested estimators.

\subsection{Ideal contrast events and the input boundary}
For rendered RGB channels normalized to $[0,1]$, the sensor uses the luminance proxy
\begin{equation}
 L=\log(0.2126R+0.7152G+0.0722B+1/255).
\end{equation}
Each pixel emits signed events when the change relative to its retained reference crosses integer multiples of contrast threshold 0.2. The reference advances by these multiples; event times are linearly interpolated between consecutive native log-intensity samples. Positive and negative events are accumulated separately in causal intervals $(t_{k-1},t_k]$ at 60\,Hz. The sensor is initialized with an empty first bin.

This is an ideal deterministic sensor: there is no radiometric linearization, threshold mismatch, shot noise, refractory period, finite analog bandwidth or readout saturation. It is neither ESIM nor v2e. Saved 60\,Hz lossless RGB videos are acquisition-quality records, not model inputs. The inference boundary permits event coordinates, polarities, timestamps and fixed geometric priors only. Ground-truth joints, activity labels, RGB, IMU, camera pose and motion commands are excluded from prediction and compensation.

\subsection{Completed-cycle ground truth}
Let $q(t)$ be the actual displacement of the designated primary wing joint from its zero position. A completed wingbeat ends at a positive-going zero crossing when the interval since the preceding positive-going crossing has visited both $q\geq0.05$ and $q\leq-0.05$\,rad and has duration no greater than 0.75\,s. Ground-truth crossings require actual positive velocity above $10^{-4}$\,rad/s and are interpolated from the high-rate physical trajectory. The amplitude and duration gates reject small oscillations and crossings separated by a long pause. Two synchronously flapping wings count as one cycle. A first crossing without a qualifying preceding cycle does not count.

All methods use the same causal complete-cycle decoder on their displacement estimates $\hat q_k$. Its crossing direction is derived from successive estimates; an auxiliary predicted velocity is not used to grant a count. Counting state is preserved across the stream. A crossing can be time-interpolated only once the current observation arrives; its estimated occurrence time and the time at which it becomes available are therefore distinct.

\subsection{Metrics and independent experimental units}
For stream $i$, let $N_i$ and $\hat N_i$ denote true and predicted completed cycles in the primary interval $(0.5,T_i]$. We report
\begin{equation}
 \mathrm{MAE}=\frac{1}{M}\sum_i|\hat N_i-N_i|,\qquad
 \mathrm{Exact}=\frac{100}{M}\sum_i\mathbf{1}[\hat N_i=N_i].
\end{equation}
MAE is measured in cycles per stream, not hertz. Exact-count accuracy can conceal cancellation between missed and extra cycles. We therefore also match individual predicted and true cycle endpoints one-to-one within $2/60=33.33$\,ms, maximizing match cardinality and then minimizing total absolute timing error. Micro-aggregated precision, recall and $F_1=2TP/(2TP+FP+FN)$ assess both overcounting and missed beats. Timing errors alone would omit missed cycles and are not the principal accuracy measure.

The primary metric retains startup failures, invalid frequency estimates and pauses; no streams or bad frames are discarded. A secondary interval $(2,T_i]$ assesses sensitivity to estimator startup. It was specified after validation and weight freezing but before any held-out predictions, and was not used for selection. The counter is not reset at 2\,s. We report this timing explicitly rather than describing the secondary analysis as an initial preregistration.

For paired ablations, errors are first averaged across the three target types and three CNN seeds within each independent test scene, separately for distance and motion condition. Differences are bootstrapped over six scenes with 5,000 draws, seed 2026092731, using percentile 95\% intervals. These are descriptive, small-sample intervals conditional on the acquisition design, without a multiple-testing correction. Repeated views and initializations are not additional independent scenes.

\section{Methods}
\begin{figure}[t]
\centering
\includegraphics[width=\linewidth]{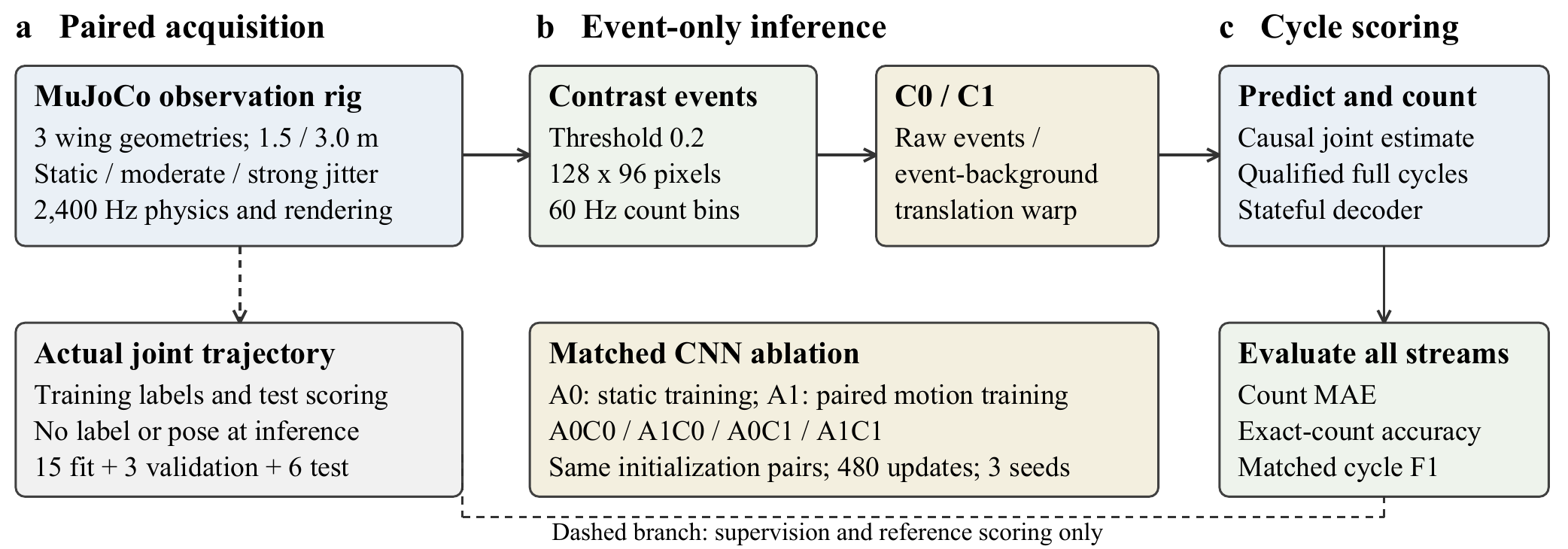}
\caption{Controlled event-only counting protocol. Solid arrows denote inference information flow; the lower branch supplies labels for fitting and scoring only. The $2\times2$ ablation changes motion training and event-background compensation while holding the temporal CNN, initialization pairing and update budget fixed. No simulator state or RGB enters the inference branch.}
\label{fig:method}
\end{figure}

\subsection{Fixed causal temporal CNN}
Each two-polarity count image is reduced to an $8\times8$ grid of spatial means per polarity plus two global means. Logarithmic compression gives 130 features. Means and scales are estimated from fitting observations only, scales are floored at 0.01, and standardized values are clipped to $[-20,20]$. The CNN has a 130-to-64 linear projection with GELU, followed by four residual temporal blocks with width 64, kernel size three and dilations 1, 2, 4 and 8. Each block uses left padding, GELU, dropout 0.05 and a pointwise mixing convolution. A linear readout predicts two values. The model has 74,562 trainable parameters and depends on the current bin and 30 preceding bins.

Training targets are $q/0.5$ and $\dot q/(2\pi\cdot5\cdot0.5)$. The objective combines their mean squared error with a weight-0.25 temporal-difference mean squared error. The velocity head is auxiliary supervision; the count decoder uses the displacement head. Models are trained independently for each target and distance. This design supports a controlled comparison at known geometry and scale, not transfer to unseen vehicle types or distances.

\subsection{Matched motion-training and compensation ablation}
Factor $A$ controls training views: $A0$ uses static observations only, whereas $A1$ draws either the static or paired moderate-motion view for each sampled base scene/window using a separate random coin. Factor $C$ controls representation: $C0$ uses raw event bins; $C1$ compensates event bins at both training and inference. All four groups use the same architecture, equal update budgets, matched base windows and the same initial weights for each target, distance and seed. The number of models is $4\times2\times3\times3=72$.

Each run completes 20 epochs of 24 updates, batch size eight, with 192-bin windows and the first 30 bins excluded from the training loss. AdamW uses learning rate 0.001 with cosine decay to 0.0001, weight decay 0.0001 and gradient-norm clipping at five. All groups are evaluated on the same six validation streams per target/distance: three scenes with both static and moderate motion. Checkpoints at epochs 0, 5, 10, 15 and 20 are selected by mean normalized event error $(FP+FN)/\max(1,N)$, then count MAE, then earlier epoch. This complete pipeline, including selection, is the object of comparison.

\subsection{Strictly event-only translation compensation}
Compensation estimates apparent background translation from an exponential activity surface formed from recent event counts. The top and bottom 16 image rows are fixed background regions; they are not simulator-provided segmentation masks. Forward and backward pyramidal Lucas--Kanade tracking on these event-derived surfaces supplies candidate displacements. Robust median consensus and confidence gates reject unreliable updates. Accepted displacements accumulate into an offset; inverse bilinear translation warps both polarity count images before feature pooling. Invalid estimates retain the preceding offset and remain in evaluation. Appendix~\ref{app:comp} lists the gates.

The module is a small-angle image-translation approximation. It does not estimate full camera pose, reproject asynchronous events individually or remove background events from the learning representation. Bilinear resampling and zero padding can change event-count mass and can move events out of the image. We consequently test its effect rather than assuming that accepted flow estimates improve counting.

\subsection{Baselines and causal period-to-count adaptation}
All baselines receive raw event observations and use the same fit/validation scene boundary. Calibration uses both static and moderate-motion fitting streams, giving classical methods access to the same motion conditions as $A1$. No test labels select hyperparameters.

\paragraph{Lagged ridge.}
A supervised ridge regressor predicts the normalized joint outputs from features at lags 0, 1, 2, 4 and 8. The regularization grid is $\{0.1,1,10,100\}$, selected on validation. Its displacement output enters the shared decoder.

\paragraph{Fourier and autocorrelation.}
A fit-only supervised linear projection maps current raw features to a displacement proxy. Its regularization is selected by validation displacement error. A fast Fourier transform (FFT) or autocorrelation (ACF) estimates frequency within 2--10\,Hz from a causal window of 0.5, 1.0 or 1.5\,s, selected by validation counting. FFT uses a Hann window and zero-padding to 2,048 samples. ACF considers lags 6--30 bins and selects the earliest positive peak reaching 90\% of the largest peak, followed by quadratic interpolation. This explicitly avoids choosing a two-period peak solely because it is larger.

At the estimated frequency, a causal least-squares sinusoid-plus-offset fit to the displacement proxy predicts the current displacement. The adapter requires at least 24 samples and recent 0.1\,s displacement range above 0.02\,rad; unavailable frequency or insufficient activity gives zero displacement. Estimates are clipped to $\pm0.8$\,rad. This shared phase reconstruction makes the rate estimators into cycle counters. The resulting baselines are calibrated, partly supervised pipelines, not unsupervised frequency-only methods.

\paragraph{EEPPR with counting adaptation.}
We retain unmodified upstream correlation and peak/template-selection functions from commit \texttt{6d45b02} of EEPPR \cite{eepprcode}. Our adapter forms one-second event windows at 1\,ms temporal resolution, uses $18\times18$ spatial patches in a fixed frontal region, updates every 0.2\,s and takes the median of valid patch frequencies in 2--10\,Hz. Its event-count parameter for template selection is selected from 60 and 180 on validation. It uses the same supervised displacement proxy and causal phase/count adapter described above. The first full EEPPR window becomes available after one second, which incurs a startup cost under the primary metric. The comparison assesses this documented adaptation, not the full original EEPPR software stack or its published frequency accuracy.

\section{Results}
\subsection{Motion training improves the matched CNN pipeline}
\begin{table}[!htbp]
\centering
\footnotesize
\caption{Near (1.5 m) held-out results in the primary interval $(0.5,T]$. MAE is in cycles (lower is better); exact-count accuracy (Ex.) and event F1 are percentages (higher is better). Each classical row aggregates 18 streams per view; each CNN row aggregates the same 18 streams over three seeds (54 predictions). The independent sample remains six scenes. A denotes motion training and C denotes event-only compensation.}\label{tab:near}
\begin{tabular}{lrrrrrrrrr}
\toprule
Method & \multicolumn{3}{c}{Static} & \multicolumn{3}{c}{Moderate motion} & \multicolumn{3}{c}{Stronger motion} \\
\cmidrule(lr){2-4}\cmidrule(lr){5-7}\cmidrule(lr){8-10}
 & MAE & Ex. & F1 & MAE & Ex. & F1 & MAE & Ex. & F1 \\
\midrule
Lagged ridge & 2.056 & 55.56 & 96.07 & 11.611 & 33.33 & 86.13 & 22.389 & 22.22 & 73.95 \\
FFT + count & 3.778 & 33.33 & 96.02 & 0.944 & 33.33 & 96.73 & 14.222 & 11.11 & 80.60 \\
ACF + count & 12.222 & 16.67 & 87.24 & 7.222 & 16.67 & 89.14 & 16.000 & 0.00 & 80.06 \\
EEPPR adapted & 9.667 & 0.00 & 89.03 & 13.889 & 0.00 & 79.59 & 29.222 & 0.00 & 58.16 \\
A0C0 & 0.241 & 77.78 & 99.71 & 31.130 & 33.33 & 69.25 & 37.593 & 18.52 & 59.17 \\
A1C0 & 0.278 & 72.22 & 99.68 & 3.185 & 59.26 & 96.63 & 6.889 & 44.44 & 85.13 \\
A0C1 & 0.241 & 75.93 & 99.68 & 33.278 & 33.33 & 65.66 & 36.741 & 16.67 & 59.18 \\
A1C1 & 0.259 & 77.78 & 99.57 & 4.630 & 50.00 & 88.65 & 10.815 & 37.04 & 79.59 \\
\bottomrule
\end{tabular}
\end{table}
\begin{table}[!htbp]
\centering
\footnotesize
\caption{Far (3.0 m) held-out results in the primary interval $(0.5,T]$. MAE is in cycles (lower is better); exact-count accuracy (Ex.) and event F1 are percentages (higher is better). Each classical row aggregates 18 streams per view; each CNN row aggregates the same 18 streams over three seeds (54 predictions). The independent sample remains six scenes. A denotes motion training and C denotes event-only compensation.}\label{tab:far}
\begin{tabular}{lrrrrrrrrr}
\toprule
Method & \multicolumn{3}{c}{Static} & \multicolumn{3}{c}{Moderate motion} & \multicolumn{3}{c}{Stronger motion} \\
\cmidrule(lr){2-4}\cmidrule(lr){5-7}\cmidrule(lr){8-10}
 & MAE & Ex. & F1 & MAE & Ex. & F1 & MAE & Ex. & F1 \\
\midrule
Lagged ridge & 4.944 & 38.89 & 92.79 & 18.611 & 22.22 & 79.28 & 38.667 & 5.56 & 58.51 \\
FFT + count & 3.611 & 38.89 & 95.79 & 7.056 & 22.22 & 89.56 & 32.333 & 11.11 & 60.67 \\
ACF + count & 11.722 & 11.11 & 85.73 & 12.833 & 11.11 & 80.66 & 34.444 & 0.00 & 54.49 \\
EEPPR adapted & 13.444 & 0.00 & 84.73 & 33.556 & 0.00 & 51.02 & 45.444 & 0.00 & 27.02 \\
A0C0 & 7.537 & 62.96 & 92.37 & 42.019 & 16.67 & 53.46 & 46.741 & 11.11 & 38.05 \\
A1C0 & 0.796 & 64.81 & 98.44 & 5.444 & 46.30 & 90.82 & 10.167 & 16.67 & 69.09 \\
A0C1 & 7.481 & 64.81 & 92.68 & 45.167 & 16.67 & 56.27 & 49.130 & 11.11 & 38.83 \\
A1C1 & 0.870 & 62.96 & 98.19 & 10.185 & 31.48 & 77.40 & 13.667 & 20.37 & 70.83 \\
\bottomrule
\end{tabular}
\end{table}
Paired motion training substantially lowered counting error under the tested jitter (Tables~\ref{tab:near} and~\ref{tab:far}). Without compensation, moderate-motion MAE decreased from 31.130 to 3.185 cycles at 1.5\,m and from 42.019 to 5.444 at 3.0\,m. Under stronger held-out jitter it decreased from 37.593 to 6.889 and from 46.741 to 10.167, respectively. Scene-paired estimates retained large uncertainty but were negative in all four conditions (Table~\ref{tab:effects}).

The benefits must be interpreted as effects on the specified fitting-and-selection pipeline. Two static-trained far/BIRD\_SIM seed-0 configurations ($A0C0$ and $A0C1$) selected epoch-zero weights. Every run completed 480 updates, but the validation rule preferred the initial near-zero outputs to later weights with excessive false events. Those checkpoints were retained without post-test replacement. Consequently, the far-distance contrast includes a model-selection failure in the static-trained controls; it does not isolate an intrinsic representation benefit among equally successful trained checkpoints. The full selection record is given in Appendix~\ref{app:selection}.

\subsection{The tested compensation does not improve count MAE}
With motion training enabled, compensation increased moderate-motion MAE from 3.185 to 4.630 cycles at 1.5\,m and from 5.444 to 10.185 at 3.0\,m. Under stronger jitter it increased MAE from 6.889 to 10.815 and from 10.167 to 13.667. These paired increases had descriptive bootstrap intervals above zero (Table~\ref{tab:effects}). Thus this compensation implementation provides no added MAE benefit in the tested moving-camera settings.

The metrics need not rank methods identically. At 3.0\,m under stronger motion, $A1C1$ has slightly higher exact-count accuracy and event F1 than $A1C0$, despite worse MAE. A smaller number of exactly counted streams can coexist with smaller large errors, and total-count equality can coexist with incorrect individual cycle times. These disagreements are the reason to report all three metrics.
\begin{table}[!htbp]
\centering
\small
\caption{Scene-paired differences in count MAE. Negative values favor the first named pipeline. Percentile 95\% intervals use 5,000 resamples of six scenes; targets and seeds are averaged within each scene before resampling.}\label{tab:effects}
\begin{tabular}{llrr}
\toprule
Condition & Comparison & MAE difference & 95\% interval \\
\midrule
Near / moderate & A1C0 $-$ A0C0 & -27.944 & [-51.056, -7.204] \\
Near / moderate & A1C1 $-$ A1C0 & 1.444 & [0.037, 3.852] \\
Near / strong & A1C0 $-$ A0C0 & -30.704 & [-49.722, -10.481] \\
Near / strong & A1C1 $-$ A1C0 & 3.926 & [1.148, 6.870] \\
Far / moderate & A1C0 $-$ A0C0 & -36.574 & [-61.456, -13.519] \\
Far / moderate & A1C1 $-$ A1C0 & 4.741 & [1.056, 8.944] \\
Far / strong & A1C0 $-$ A0C0 & -36.574 & [-58.037, -16.402] \\
Far / strong & A1C1 $-$ A1C0 & 3.500 & [0.352, 7.426] \\
\bottomrule
\end{tabular}
\end{table}

\subsection{A classical baseline wins one important condition}
The Fourier baseline attained 0.944 cycles MAE at 1.5\,m under moderate motion, compared with 3.185 for $A1C0$, while their event F1 scores were 96.73\% and 96.63\%. Conversely, $A1C0$ had higher exact-count accuracy (59.26\% versus 33.33\%). The neural model therefore did not uniformly dominate a simple calibrated periodic estimator. At the far distance and under stronger motion, the Fourier pipeline was less accurate in MAE than $A1C0$. EEPPR with the present counting adapter had larger count errors than $A1C0$ in every aggregate distance/view setting, but these results cannot be attributed solely to the upstream frequency estimator: region selection, temporal quantization, missing estimates and phase reconstruction all contribute.

Startup partly affected periodic baselines, without removing the main pattern. On the secondary post-2\,s interval, near moderate-motion FFT MAE was 0.889 and $A1C0$ MAE was 3.148. Near static EEPPR-adapted MAE decreased from 9.667 to 6.556; under far moderate motion it decreased from 33.556 to 28.500 and remained high. Table~\ref{tab:startup} reports the interval sensitivity for representative methods. No hyperparameter was changed for this analysis.
\begin{table}[!htbp]
\centering
\small
\caption{Startup sensitivity: primary $(0.5,T]$ versus secondary $(2,T]$ count MAE, with continuous decoder state and unchanged selected models. The secondary interval was fixed before test prediction but after validation.}\label{tab:startup}
\begin{tabular}{lrrrrrr}
\toprule
 & \multicolumn{2}{c}{FFT + count} & \multicolumn{2}{c}{EEPPR adapted} & \multicolumn{2}{c}{A1C0} \\
Condition & Primary & Post-2 s & Primary & Post-2 s & Primary & Post-2 s \\
\midrule
Near / static & 3.778 & 3.056 & 9.667 & 6.556 & 0.278 & 0.333 \\
Near / moderate & 0.944 & 0.889 & 13.889 & 10.056 & 3.185 & 3.148 \\
Near / strong & 14.222 & 12.167 & 29.222 & 24.833 & 6.889 & 6.352 \\
Far / static & 3.611 & 2.389 & 13.444 & 9.667 & 0.796 & 0.778 \\
Far / moderate & 7.056 & 5.944 & 33.556 & 28.500 & 5.444 & 5.426 \\
Far / strong & 32.333 & 29.056 & 45.444 & 40.056 & 10.167 & 9.278 \\
\bottomrule
\end{tabular}
\end{table}

\subsection{Background-dependent failures and compensation availability}
\begin{table}[!htbp]
\centering
\small
\caption{Descriptive background subgroups: count MAE under camera motion. Each background contains only three independent test scenes. CNN errors average the three seeds.}\label{tab:background}
\begin{tabular}{llrrrr}
\toprule
Condition & Background & FFT + count & A0C0 & A1C0 & A1C1 \\
\midrule
Near / moderate & plain & 1.111 & 0.778 & 0.630 & 0.630 \\
Near / moderate & textured & 0.778 & 61.481 & 5.741 & 8.630 \\
Near / strong & plain & 13.889 & 4.815 & 0.370 & 0.370 \\
Near / strong & textured & 14.556 & 70.370 & 13.407 & 21.259 \\
Far / moderate & plain & 10.000 & 8.815 & 2.259 & 2.000 \\
Far / moderate & textured & 4.111 & 75.222 & 8.630 & 18.370 \\
Far / strong & plain & 22.000 & 18.000 & 5.741 & 6.111 \\
Far / strong & textured & 42.667 & 75.481 & 14.593 & 21.222 \\
\bottomrule
\end{tabular}
\end{table}
Plain and textured conditions must be distinguished when interpreting motion interference (Table~\ref{tab:background}). The split contains only three test scenes of each background, so these are descriptive subgroups, not independently powered confirmation experiments. Accepted compensation updates were absent in static and plain-background views. On textured moderate-motion test streams, accepted-update fractions were approximately 92.9\% near and 98.6\% far; the accumulated offset reached its eight-pixel bound. Textured strong-motion streams had approximately 99.8\% accepted updates. These values measure the algorithm's internal availability, not pose accuracy. High acceptance did not guarantee improved counting.

\subsection{Integrity checks and preserved failures}
Raw-event reaggregation was checked for all 324 streams, and actual-joint cycle labels were reconstructed independently in code. Paired physics arrays were equal across distances and camera views. Audits checked fitting-only normalization, matching initialization and sampled base windows, 72 selection records and six baseline calibrations. All 1,728 held-out predictions (1,296 CNN and 432 baseline predictions) passed separately implemented cycle reconstruction, event matching and causal count-drift checks. Prefix/chunk consistency tests used fitting observations and a $3\times10^{-5}$ output tolerance. These checks test implementation consistency; they are not third-party human validation or evidence that the sensor model matches hardware.

Method code and all chosen weights and baseline settings were frozen before held-out prediction. The acquisition design was recorded before capture, while its source-hash record was made early after recording began. A preflight import-order repair and an ACF peak-selection correction preceded training and final method freezing. The work was not externally preregistered. This chronology is retained to make the degree of prospective control clear.

\section{Discussion}
The strongest supported conclusion is conditional: motion examples improved complete-wingbeat counting for a fixed CNN and fixed training budget in this paired synthetic protocol. The result does not require a new neural architecture. It also does not show that background compensation is unnecessary in general. A translation warp may be too restrictive for angular image motion, its accumulated estimate can drift or saturate, and interpolation can alter sparse event structure. These are plausible explanations for the adverse result, not measured causal mechanisms; a pose-error audit, richer motion model and interpolation controls would be needed to separate them.

The Fourier result argues for retaining classical periodic estimators as serious baselines. In a constrained, low-frequency task with calibrated geometry and supervised phase reconstruction, a small estimator can outperform a neural model on aggregate count error. Equally, an aggregate count alone can hide when errors occur. Reporting cycle-time matching, count MAE and exact totals avoids treating frequency accuracy or a single favorable score as successful counting.

Several boundaries limit external validity. First, only six independent test scenes were used, and bootstrap resampling cannot create new background or motion diversity. Second, models are target- and distance-specific; three geometries do not establish open-world recognition. Third, the sensor lacks realistic noise, latency, threshold variation and bandwidth limits. No render-rate convergence study or real event-camera experiment is reported. Fourth, the cameras remain front-facing and undergo a narrow angular-jitter family; translation, occlusion, lateral targets and aggressive flight are absent. Fifth, the test rig uses externally maintained poses and prescribed wing trajectories, with no validated aerodynamic control. Sixth, two epoch-zero controls expose sensitivity to selection and make some far-distance gains partly a comparison against failed selected models. Finally, inference energy, hardware latency and compute-matched exhaustive baseline searches were not measured. Simulated event-time confirmation is not wall-clock processing latency.

The next evidential steps are therefore specific: increase independent scenes and background diversity with a fresh held-out split, replace the ideal sensor with a calibrated or measured event stream, assess compensation against independent pose ground truth for analysis only, and evaluate on synchronized physical wing-joint or high-speed-video references. A broader architecture comparison would require retraining SNN and transformer models under this same frozen protocol rather than importing accuracy values from the earlier RGB study.

\section{Conclusion}
This controlled MuJoCo benchmark distinguishes completed-cycle counting from periodic-rate estimation and isolates motion training from strictly event-only translation compensation. Paired motion training improved a fixed CNN under moderate and stronger angular jitter, while the tested compensation worsened count MAE. A Fourier baseline was best in near moderate-motion MAE. The evidence supports careful metric design, matched ablations and transparent baseline adaptation; it remains limited to a small ideal-sensor simulation and does not establish real-flight performance.

\section*{Data, code and assistance statement}
Acquisition manifests, source hashes, trained weights, validation selections, per-stream predictions and audit outputs are retained with the experiment. The manuscript source contains the numerical tables and figure needed to reproduce this document. A public raw-data or code repository has not been deposited for this version, and no public accession is claimed. Upstream geometry sources are cited below; their redistribution rights remain with their owners. The adapted EEPPR core retains its GPL-3.0 license in the local experiment package. AI assistance was used for implementation, analysis checks, figure layout and English drafting. The reported numerical results are taken from executed simulations and frozen evaluation artifacts, not generated example values. The schematic figure is deterministic vector artwork and is not an experimental observation.

\appendix
\section{Geometry provenance and adaptations}\label{app:assets}
The observer model is adapted from Google DeepMind's MuJoCo Menagerie Crazyflie 2 model (\url{https://github.com/google-deepmind/mujoco_menagerie}). Target sources and fixed revisions are:
\begin{itemize}
\item RLFlapping: \url{https://github.com/Mostafa12d/RLFlapping}.\\Commit \texttt{3577d0c41d3d40e1f82ecdcc55e490e33a943c4f};\\static wingspan 1.300\,m; primary joints \texttt{J\_flap\_L} and \texttt{J\_flap\_R}.
\item flappy\_v2: \url{https://github.com/JiazeCai/flappy_v2}.\\Commit \texttt{d050e8b5c4f46beec7088189efd668485e498ccc}; static wingspan 0.405\,m; primary joints \texttt{J5} and \texttt{J5R}.
\item BIRD\_SIM: \url{https://github.com/Milkomedia/BIRD_SIM}.\\Commit \texttt{c5d5c3c123c0d648bed07205a5985725f8e07217}; static wingspan 0.750\,m; primary joints \texttt{J1} and \texttt{J7}.
\end{itemize}
The imported joint hierarchy, axes, inertias and mesh scales were retained. Original scene decorations and auxiliary thrusters were removed; free root joints and observation-rig tracking were added. BIRD\_SIM's aerodynamic/motor implementation and the source projects' learned policies were not transferred. The flappy\_v2 MIT license was retained. No license file was found in the pinned RLFlapping and BIRD\_SIM trees during asset preparation; this manuscript does not assign an open-source license to those assets or redistribute their meshes in the manuscript source archive.

\section{Nonstationary schedules and sampling}\label{app:schedule}
For a chirp of duration $T$, frequency increases linearly from 3 to 7\,Hz on $[0,T/2]$ and decreases linearly to 3\,Hz on $[T/2,T]$. Phase is the analytic integral of frequency, not $f(t)t$. The stop--restart profile has active intervals $[0,0.20T)$ at 3\,Hz, $[0.35T,0.60T)$ at 5\,Hz and $[0.75T,T]$ at 7\,Hz. Phase freezes in the two rest intervals. A quintic smoothstep $6u^5-15u^4+10u^3$ ramps amplitude down and up over 0.4\,s at stop/restart boundaries; the final active segment continues beyond the recording endpoint. Initial phases and amplitudes are sampled by the scene configuration. Labels are derived from actual tracked motion, not the analytic phase. Each recording follows a 0.8\,s tracking warm-up; the event sensor is then initialized for the recorded stream.
Scene seeds are generated as \texttt{SeedSequence([2026092719, i])} for scene index $i$. Duration is 12\,s for even $i$ and 16\,s for odd $i$; profile is selected by $i\bmod3$. Checker texture is assigned when $(\lfloor i/3\rfloor+i\bmod3)\bmod2=1$. Geometry and excitation are paired between distances and views. Model inputs use only the resulting events, not these generation variables.

\section{Compensation and EEPPR implementation details}\label{app:comp}
The activity surface uses $S_k=0.65S_{k-1}+0.35\log(1+E^+_k+E^-_k)$, followed by a $3\times3$ Gaussian filter with standard deviation 0.6. Background regions are rows 0--15 and 80--95, with the background 99th percentile mapped to intensity 180 for tracking. Tracking requires at least 20 active background pixels above 0.05. Corner detection uses up to 100 points, quality 0.04, minimum separation four pixels and block size three. Lucas--Kanade uses $9\times9$ windows, one pyramid level, 20 iterations and termination epsilon 0.01. Forward/backward error must be below 0.5 pixels. At least eight tracks must survive; residual consensus uses threshold $\max(0.3,2.5\,\mathrm{median\ residual})$, at least eight inliers and an inlier fraction of 60\%. Per-bin shift magnitude cannot exceed 2.5 pixels; cumulative offsets are clipped to $\pm8$ pixels per axis. Failed gates retain the prior offset rather than dropping the observation.

The EEPPR counting adapter uses ROI $x=10,\ldots,117$ and $y=21,\ldots,74$, giving 18 non-overlapping $18\times18$ patches. Quantized duplicate event coordinates are collapsed to Boolean occupancy; competing polarities in a cell retain the last polarity. Upstream functions \texttt{correlate\_3d}, \texttt{find\_periodic\_peaks} and \texttt{find\_template\_depth} are loaded without modifying their source. The adapter avoids the upstream file-format dependency stack and evaluates the documented fixed ROI rather than interactive per-test cropping. These engineering choices are part of the adapted baseline and restrict conclusions about the original method.

\Needspace{16\baselineskip}
\section{All checkpoint selections}\label{app:selection}
Each cell lists the selected epochs for seeds 0, 1 and 2. Epoch zero is an initialization checkpoint, not a trained checkpoint. All runs nevertheless executed the same 480 training updates. The table exposes this failure instead of silently replacing selected models after testing.
\begin{table}[!htbp]
\centering
\small
\caption{Selected epochs for all 72 CNN runs (seed order 0, 1, 2).}\label{tab:selection}
\begin{tabular}{llrrrr}
\toprule
Distance & Target & A0C0 & A1C0 & A0C1 & A1C1 \\
\midrule
Near & rlflapping & 20, 15, 20 & 10, 20, 15 & 10, 15, 15 & 20, 20, 10 \\
Near & flappy\_v2 & 10, 5, 15 & 15, 15, 20 & 20, 10, 10 & 10, 20, 10 \\
Near & bird\_sim & 5, 5, 5 & 20, 20, 20 & 10, 5, 5 & 20, 20, 10 \\
Far & rlflapping & 5, 5, 5 & 20, 15, 10 & 5, 5, 5 & 20, 20, 20 \\
Far & flappy\_v2 & 20, 5, 20 & 20, 20, 15 & 20, 10, 20 & 20, 20, 20 \\
Far & bird\_sim & 0, 10, 5 & 20, 20, 20 & 0, 15, 5 & 20, 20, 20 \\
\bottomrule
\end{tabular}
\end{table}

\paragraph{Artifact provenance.}
Source, input and model hashes and the test chronology are preserved separately. Tables are generated from frozen results without retraining during writing. The local review package contains code, weights, selections, predictions and audits, but excludes raw acquisitions. Manuscript edits do not modify experimental records.

\begin{thebibliography}{10}
\bibitem{frequency} B. Pfrommer. Frequency Cam: Imaging Periodic Signals in Real-Time. \emph{arXiv:2211.00198}, 2022. \url{https://arxiv.org/abs/2211.00198}.
\bibitem{eeppr} J. Kol\'{a}\v{r}, R. \v{S}petl\'{i}k, and J. Matas. EEPPR: Event-based Estimation of Periodic Phenomena Rate using Correlation in 3D. \emph{arXiv:2408.06899}, 2024. \url{https://arxiv.org/abs/2408.06899}.
\bibitem{repnet} D. Dwibedi, Y. Aytar, J. Tompson, P. Sermanet, and A. Zisserman. Counting Out Time: Class Agnostic Video Repetition Counting in the Wild. In \emph{CVPR}, pp. 10387--10396, 2020. \url{https://arxiv.org/abs/2006.15418}.
\bibitem{tcn} S. Bai, J. Z. Kolter, and V. Koltun. An Empirical Evaluation of Generic Convolutional and Recurrent Networks for Sequence Modeling. \emph{arXiv:1803.01271}, 2018. \url{https://arxiv.org/abs/1803.01271}.
\bibitem{mujoco} Google DeepMind. MuJoCo: Advanced Physics Simulation. Software and documentation. \url{https://mujoco.org/}. Accessed September 2026.
\bibitem{esim} H. Rebecq, D. Gehrig, and D. Scaramuzza. ESIM: an Open Event Camera Simulator. In \emph{Conference on Robot Learning}, PMLR 87, pp. 969--982, 2018. \url{https://proceedings.mlr.press/v87/rebecq18a.html}.
\bibitem{v2e} Y. Hu, S.-C. Liu, and T. Delbruck. v2e: From Video Frames to Realistic DVS Events. \emph{CVPR Workshops}, 2021. \url{https://arxiv.org/abs/2006.07722}.
\bibitem{prior} Zhang Nengbo. Optical-Flow Wingbeat Counting in MuJoCo: A Comparison of Convolutional, Spiking, and Attention-Based Temporal Models. \emph{arXiv:2609.17308}, 2026. \url{https://arxiv.org/abs/2609.17308}.
\bibitem{eepprcode} J. Kol\'{a}\v{r}, R. \v{S}petl\'{i}k, and J. Matas. EEPPR source code.\\Commit \texttt{6d45b02ba6d61a9515e2dbe6189637f6a740ea44}. \url{https://github.com/JackPieCZ/EEPPR}. GPL-3.0.
\end{thebibliography}
\end{document}